\documentclass[10pt,twocolumn,letterpaper]{article}
\usepackage[pagenumbers]{cvpr}              % To produce the CAMERA-READY version
\usepackage{multirow}
\usepackage{indentfirst} 
\usepackage{pifont}
\usepackage{adjustbox}
\usepackage{bm}
\usepackage{wrapfig, blindtext} 
\usepackage{multirow}
\usepackage{bbding}
\usepackage{makecell}
\definecolor{cvprblue}{rgb}{0.21,0.49,0.74}
\usepackage[pagebackref,breaklinks,colorlinks,allcolors=cvprblue]{hyperref}
\def\paperID{5396} % *** Enter the Paper ID here
\def\confName{CVPR}
\def\confYear{2026}
\title{It Takes Two: A Duet of Periodicity and Directionality for Burst Flicker Removal}
\author{Lishen Qu$^{1,3,5}$, Shihao Zhou$^{1,3}$ Jie Liang$^{5}$, Hui Zeng$^{5}$, Lei Zhang$^{4,5}$, Jufeng Yang$^{1,2,3,}$\thanks{Corresponding Author.} \\
{$^{1}$Nankai International Advanced Research Institute (SHENZHEN·FUTIAN)} \\
{$^{2}$Peng Cheng Laboratory \qquad $^{3}$College of Computer Science, Nankai University} \\
{$^{4}$The Hong Kong Polytechnic University \qquad $^{5}$OPPO Research Institute}\\
{\tt\small \ \{qulishen,zhoushihao96\}@mail.nankai.edu.cn, \ liang27jie@163.com},   \\
{\tt\small  \ cshzeng@gmail.com,\ cslzhang@comp.polyu.edu.hk, \ yangjufeng@nankai.edu.cn} \ \\
}
\begin{document}
\maketitle
\begin{abstract}
Flicker artifacts, arising from unstable illumination and row-wise exposure inconsistencies, pose a significant challenge in short-exposure photography, severely degrading image quality.
Unlike typical artifacts, e.g., noise and low-light, flicker is a structured degradation with specific spatial-temporal patterns, which are not accounted for in current generic restoration frameworks, leading to suboptimal flicker suppression and ghosting artifacts.
In this work, we reveal that flicker artifacts exhibit two intrinsic characteristics, periodicity and directionality, and propose Flickerformer, a transformer-based architecture that effectively removes flicker without introducing ghosting.
Specifically, Flickerformer comprises three key components: a phase-based fusion module (PFM), an autocorrelation feed-forward network (AFFN), and a wavelet-based directional attention module (WDAM).
Based on the periodicity, PFM performs inter-frame phase correlation to adaptively aggregate burst features, while AFFN exploits intra-frame structural regularities through autocorrelation, jointly enhancing the network’s ability to perceive spatially recurring patterns.
Moreover, motivated by the directionality of flicker artifacts, WDAM leverages high-frequency variations in the wavelet domain to guide the restoration of low-frequency dark regions, yielding precise localization of flicker artifacts.
Extensive experiments demonstrate that Flickerformer outperforms state-of-the-art approaches in both quantitative metrics and visual quality.
The source code is available at \url{https://github.com/qulishen/Flickerformer}.
\end{abstract}    
\section{Introduction}
\label{sec:intro}
The acquisition of images under artificial light sources powered by alternating current (AC) often leads to flicker artifacts~\cite{sheinin2017computational}, posing a persistent challenge in photography.
Since the intensity of these light sources oscillates with the AC frequency, the illumination varies periodically within each cycle~\cite{yoo2014flicker,camera_sys}.
\begin{figure}[t]
    \centering
    \includegraphics[width=\linewidth]{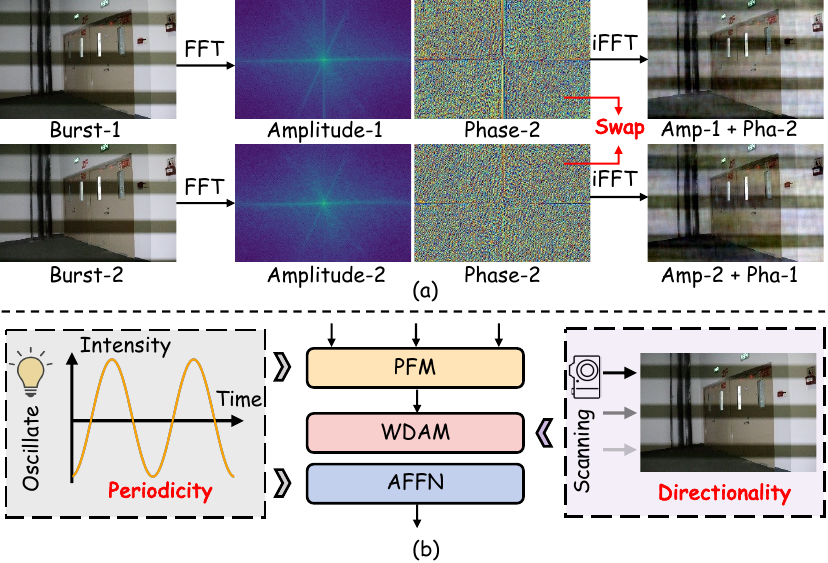}
    \caption{Motivation of Flickerformer. (a) Swapping the phase components between two consecutive flickering images leads to an exchange of flicker patterns, indicating that phase encodes the spatial distribution of flicker. (b) Illustration of the intrinsic flicker characteristics and the corresponding module designs: PFM and AFFN are devised based on periodicity, while WDAM is inspired by directionality.
}
    \label{fig:motivation}
    \vspace{-1mm}
\end{figure}
When a camera captures a frame with a short exposure time, it often covers only a fraction of an illumination cycle, resulting in a recorded image that reflects an incomplete light waveform~\cite{kim2013reducing}.
Moreover, modern cameras, which capture images using a rolling-shutter mechanism, expose the sensor line by line~\cite{rolling,vision_sensor1,vision_sensor3}, which leads to slight differences in the exposure times of different rows.
The combination of this inter-row timing difference and the oscillating illumination results in striped brightness patterns along the scanning direction~\cite{Lin,burstdeflicker}, as shown in~\cref{fig:motivation}.
Such flicker artifacts not only degrade the perceptual quality of captured images but also impair the performance of downstream vision tasks~\cite{flicker_jiankong,flicker_driving,flicker_surgery}.
Additionally, short exposure strategies are necessary in various tasks, including high dynamic range (HDR) imaging~\cite{hdr_iccv_2025_1}, slow-motion video~\cite{slow_motion}, and motion capture~\cite{motion_capture1}.
Therefore, there is an increasing need for developing a stable and generalizable solution for flicker removal.

Traditional flicker removal methods~\cite{related1,prior_flicker2,auto_flicker} attempted to suppress flicker through pattern matching or brightness approximation, yet their effectiveness remains limited.
For example, some methods~\cite{prior_flicker1,related1} exploited the periodic nature of AC-powered lighting, estimating the illumination modulation curve from the difference between the short and long exposure frames.
In addition, several hardware-based solutions~\cite{prior_flicker1,auto_flicker,reno2017powerline_embedding} tried to suppress flicker during image acquisition by integrating modulation detection or compensation mechanisms at the sensor level.
However, these methods often rely on specialized hardware designs, limiting their applicability to diverse imaging devices and broader real-world scenarios.

With the success of deep learning~\cite{resnet,vaswani2017attention,huang1}, several data-driven solutions have been proposed to tackle the flicker removal problem. Lin~\etal~\cite{Lin} introduced the first learning-based approach by synthesizing flickering images from clean ones and training a CycleGAN~\cite{cyclegan} for flicker suppression. 
Zhu~\etal~\cite{zhu2025rifle} designed a dataset synthesis scheme specifically for removing flicker artifacts caused by PWM-modulated screens.
More recently, Qu~\etal~\cite{burstdeflicker} established the first burst flicker removal benchmark, BurstDeflicker, demonstrating the potential of multi-frame restoration methods for flicker removal.
However, existing methods primarily treat flicker removal as a generic image restoration task, overlooking the underlying physical priors. 
As a result, these models often meet challenges to capture the structured nature of flicker artifacts, leading to suboptimal restoration performance, especially under serious and covert flicker conditions.

In this work, we bridge the gap between physics-based modeling and deep learning by embedding flicker priors into a neural network framework. 
As shown in~\cref{fig:motivation}, flickering images exhibit distinct periodic patterns, and swapping their phase components alters the spatial distribution of flicker across frames, which indicates that the phase information of flicker encodes its spatial distribution.
Motivated by this observation, we introduce a phase-based fusion module (PFM) and an autocorrelation feed-forward network (AFFN).
Phase correlation, a classic technique in signal processing~\cite{phase,kuglin1975phase}, effectively measures cyclic or translational similarity between images in the frequency domain.
Our PFM leverages this property to align and fuse multi-frame features, capturing inter-frame variations and effectively extracting useful features of the reference frames.
After feature fusion, the AFFN models intra-frame periodic structures through autocorrelation~\cite{auto}, which provides a principled way to detect repeating patterns within a signal.
By jointly exploiting inter-frame phase correlation and intra-frame autocorrelation, our framework effectively leverages the periodicity of flicker, yielding more stable and coherent restoration results.

Besides, flicker artifacts exhibit strong directionality due to the rolling-shutter scanning mechanism of modern image sensors~\cite{rolling}.
As shown in~\cref{fig:motivation}, these artifacts typically appear as horizontally or vertically aligned stripes, producing structured high-frequency luminance oscillations and low-frequency dark bands along the scanning direction.
To exploit this property, we propose a wavelet-based directional attention module (WDAM), which enhances the network’s precision in locating flicker regions and improves its restoration capability.
Unlike conventional convolution~\cite{cao2022swin} and self-attention~\cite{liang2021swinir,wang2022uformer,blur-3}, which process features isotropically, wavelet decomposition separates images into orientation-specific subbands.
The WDAM applies Haar wavelet~\cite{huang2017wavelet} decomposition to separate the feature into low- and high-frequency components.
This process produces orientation-specific high-frequency subbands, which naturally correspond to flicker variations.
We leverage these subbands to guide attention in the low-frequency branch for restoring flicker-affected dark regions.
By combining a dual-branch design with directional decomposition, WDAM enhances the robustness of flicker removal while reducing computational overhead.
Finally, we integrate PFM, AFFN, and WDAM into a unified transformer framework, termed Flickerformer, which jointly models periodicity-aware and direction-aware representations for effective burst flicker removal.

Our main contributions are summarized as follows:
\begin{itemize}
    \item We propose Flickerformer, a transformer-based framework designed for burst flicker removal. It achieves high-quality restoration of flickering images without introducing ghosting artifacts.
    \item Guided by the periodicity, we introduce PFM and AFFN to model inter-frame similarity and intra-frame periodic structures, respectively.
    To further exploit directionality, WDAM enhances the restoration by locating and restoring flickering regions in the wavelet domain.
    \item Extensive experiments on real-world flicker datasets demonstrate that our method consistently outperforms previous state-of-the-art approaches in both quantitative results and visual quality.
\end{itemize}

\section{Related Work}
\label{sec:related}
\noindent\textbf{Vision Transformers.}
Transformers~\cite{vaswani2017attention,huang2023iddr,huang2026nerf} have revolutionized various vision tasks by modeling long-range dependencies through self-attention.
The Vision Transformer (ViT)~\cite{dosovitskiy2020image} first demonstrated that pure transformer architectures can outperform convolutional networks when trained on large-scale datasets.
Since the computational complexity of Transformers scales quadratically with image resolution, various adaptations have been proposed in the image restoration works to alleviate this cost and make Transformers more practical for high-resolution inputs.
Uformer-based model~\cite{wang2022uformer,zhou2024seeing} adopts window-based attention to enhance local feature modeling, while SwinIR~\cite{liang2021swinir} introduces a shifting mechanism to enable richer cross-window interactions.
Restormer~\cite{zamir2022restormer} further reduces computational complexity by performing attention along the channel dimension.
In burst or video restoration, transformers also have shown strong potential in handling spatial-temporal correlations~\cite{dudhane2023burstormer,liang2024vrt, hdrtransformer}.
However, conventional attention mechanisms tend to perform implicit low-pass filtering~\cite{pass_filter}, which weakens their ability to model structured high-frequency degradations such as flicker.
In this work, we propose a WDAM that separately models low-frequency and high-frequency information, and leverages directional features in the high-frequency components to guide the restoration of low-frequency regions.

\noindent\textbf{Burst Image Restoration.}
Burst photography~\cite{burst-photo} in handheld cameras leverages multiple frames to enhance image quality under challenging conditions such as low light~\cite{burst-low-light1,burst-low-light2,burst-low-light3}, low resolution~\cite{burst-SR1,dudhane2024burst,BurstSR}, and severe noise~\cite{burst-denoise1,burst-denoise3,mildenhall2018burst}.
Traditional pipelines~\cite{mildenhall2018burst, aittala2018burst} usually involve explicit alignment, such as optical flow or patch-based matching, followed by pixel-level fusion.
While these methods improve signal-to-noise ratio and preserve fine details, they are highly sensitive to motion and tend to produce ghosting artifacts in dynamic scenes. 
To overcome these limitations, recent learning-based methods jointly perform alignment and fusion within a deep network. 
For instance, Akshay~\etal~\cite{dudhane2023burstormer} proposed Burstormer, which adopted a multi-scale hierarchical transformer, where offset features are estimated at different scales to guide feature alignment. 
Similarly, Wei~\etal~\cite{towards-burstsr} introduced FBANet, which integrated homography alignment with a federated affinity fusion mechanism, thereby improving the performance of multi-frame alignment and fusion.
Recently, diffusion-based networks~\cite{burstsr-2025-1} and Mamba-based networks~\cite{burstsr-2025-2} for burst super-resolution also demonstrated notable performance improvements.
These approaches typically assume spatially homogeneous degradations in the image, which is valid for those captured under low-resolution or low-light conditions.
However, this assumption does not hold for flickering images, which introduces non-uniform, structured periodic intensity fluctuations that vary over time.

\noindent\textbf{Flicker Removal.}
Classical methods~\cite{park2009method,camera_sys} relied on hardware-based sensors that detect flickering light sources and dynamically adjust the exposure time to mitigate flicker. 
However, simply extending the exposure time often introduces motion blur~\cite{blur-1,blur-2,blur-3}, which limits their applicability in dynamic scenes. 
Other approaches~\cite{prior_flicker1,prior_flicker2,prior_flicker3} assumed prior knowledge of the lighting system parameters and exploit this information for flicker correction, achieving satisfactory results in controlled environments but struggling in wild scenarios. 
Recent advances in deep learning have enabled significant progress in image restoration~\cite{image-restoration-1,image-restoration-2,image-restoration-3}.
Lin \etal~\cite{Lin} introduced DeflickerCycleGAN, the first data-driven approach for flicker removal, demonstrating the potential of deep neural networks for this task.
More recently, Qu~\etal~\cite{burstdeflicker} proposed the first multi-frame flicker removal dataset and built a comprehensive benchmark on several representative restoration networks~\cite{zamir2022restormer,hdrtransformer,dudhane2023burstormer}.
However, these generic restoration networks are not specifically designed for burst flicker removal, which limits their ability to capture the intrinsic characteristics of flicker. 
This paper represents the first attempt to explicitly embed flicker priors into a transformer-based architecture, thereby enhancing the robustness of flicker removal and mitigating ghosting artifacts in burst flicker removal.

\section{Proposed Method}
\label{sec:method}

\begin{figure*}[t]
    \centering
    \includegraphics[width=\textwidth]{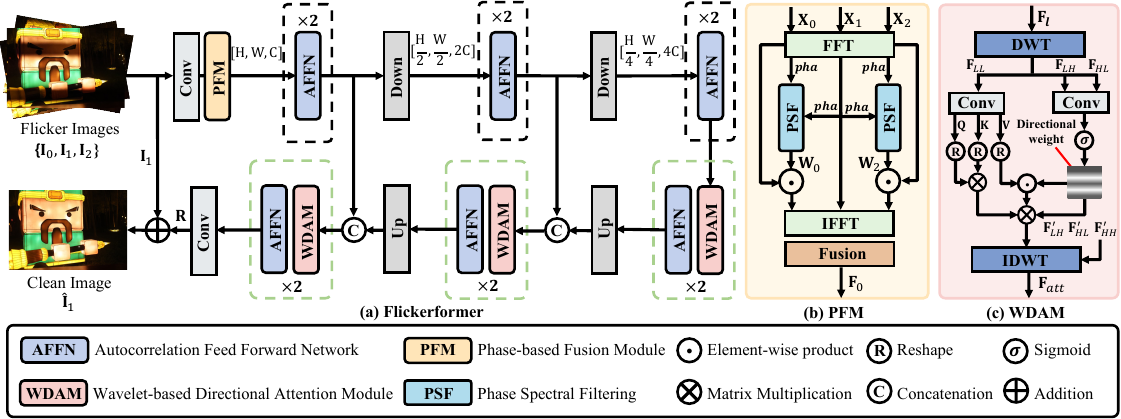}
    \caption{Overview of Flickerformer.
(a) The proposed Flickerformer adopts an asymmetric U-shaped encoder–decoder architecture. Following the previous work~\cite{burstdeflicker}, the input consists of three frames, while the output is a single restored frame.
(b) The proposed phase fusion module (PFM), which performs feature fusion based on phase correlation. The operation of the phase spectral filtering (PSF) corresponds to~\cref{equal:pf1,equal:pf2}, and the fusion module corresponds to~\cref{equal:fusion}.
(c) The proposed wavelet-based directional attention module (WDAM), including a low-frequency attention branch and a high-frequency branch to capture directional features.}
    \label{fig:pipeline}
\end{figure*}

Our goal is to restore high-quality flicker-free images by modeling the periodic degradation patterns introduced by alternating current (AC) lighting, as well as leveraging directional contextual information in the entire image. 
To this end, we propose Flickerformer, a novel transformer-based architecture specifically designed for burst flicker removal. 
Flickerformer is built upon three core components: (1) the phase fusion module (PFM) (see~\cref{fig:pipeline} (b)) and (2) the autocorrelation feed-forward network (AFFN) (see~\cref{fig:AFFN}), which exploit flicker periodicity in the frequency domain, and (3) the wavelet-based directional attention module (WDAM) (see~\cref{fig:pipeline} (c)), which captures directional characteristics of flicker in the spatial domain.

\subsection{Overall Pipeline}
\label{sec:pipeline}
The overall architecture of Flickerformer is illustrated in~\cref{fig:pipeline}. Given a base frame $\mathbf{I}_1$ and two reference frames $\mathbf{I}_0$, $\mathbf{I}_2$, forming a burst of three flickering frames with spatial resolution $H \times W \times 3$. 
We first concatenate them along the channel dimension and apply a $3\times3$ group convolution layer to extract their initial low-level features ${\mathbf{X}_t} \in \mathbb{R}^{H \times W \times C}$ independently, where $t \in \{0, 1, 2\}$ denotes the frame index in the burst sequence.
Then, the extracted features ${\mathbf{X}_t}$ are fed into the PFM for feature fusion, producing the fused low-level feature $\mathbf{F}_0 \in \mathbb{R}^{H \times W \times C}$.
Subsequently, the features $\mathbf{F}_0$ are fed into a U-shaped encoder-decoder backbone. The encoder consists of three hierarchical stages. Each stage includes multiple Transformer blocks, and the number of blocks increases with depth. 
The $l$-th encoder stage outputs a downsampled feature $\mathbf{F}_l \in \mathbb{R}^{\frac{H}{2^l} \times \frac{W}{2^l} \times 2^lC}$. 
We employ the AFFN to enhance informative representations for feature refinement.
In the decoder, we employ the WDAM, which produces the feature $\mathbf{F}_{att} \in \mathbb{R}^{\frac{H}{2^l} \times \frac{W}{2^l} \times 2^lC}$.
To be specific, the output of the $l$-th decoder and the input of the $l$-th encoder are concatenated and then processed by a $1\times1$ convolutional layer to form the input for the next module.
After upsampling to the original resolution, the final feature is passed through a $3\times3$ convolution layer to predict a residual map $\mathbf{R} \in \mathbb{R}^{H \times W \times 3}$. 
The output image is then obtained by $\hat{\mathbf{I}}_1 = \mathbf{I}_1 + \mathbf{R}$, which is the flicker-free image of the base frame.

\subsection{Frequency-Domain Periodicity Modeling}
\label{sec:periodicity}
To effectively suppress flicker artifacts, we explore the intrinsic frequency-domain characteristics of flickering images. 
As analyzed in~\cref{fig:motivation}, the flicker artifacts exhibit strong periodicity, which can be explicitly captured through frequency-phase representations. 
Accordingly, we design two complementary components that exploit this property: (1) the PFM for inter-frame fusion, and (2) the AFFN for intra-frame periodicity enhancement.

\noindent\textbf{Phase-based Fusion Module.}
Let $\mathbf{X}_t \in \mathbb{R}^{H\times W \times C}$ denote the low-level feature of the $t$-th frames.
We apply the Fast Fourier Transform (FFT) to obtain the frequency feature, which is represented by:
\begin{equation}
\tilde{\mathbf{X}}_t = \mathcal{F}(\mathbf{X}_t) 
= A_t(\mathbf{k}) e^{i\Phi_t(\mathbf{k})}, 
\quad t \in \{0,1,2\},
\end{equation}
where $i$ is the imaginary unit and $\mathbf{k}$ represents the frequency coordinates.
$\mathcal{F}(\cdot)$ is the 2D fast Fourier transform (FFT) operation. 
$A_t(\mathbf{k})$ and $\Phi_t(\mathbf{k})$ are the amplitude and phase spectra values at the $\mathbf{k}$, respectively. 

The phase spectrum has been widely recognized to capture structural and alignment information of images~\cite{kuglin1975phase}.
Since the flicker distribution primarily lies in the phase, as discussed in~\cref{fig:motivation}, we adopt phase correlation~\cite{science_phase, phase} to evaluate the similarity between the base frame and the two reference frames:
\begin{equation}
\mathbf{S}_t(\mathbf{k}) =
\Big| e^{i\Phi_{t}(\mathbf{k})} 
 \odot e^{-i\Phi_1(\mathbf{k})}\Big|, 
\quad t \in \{0,2\}.
\label{equal:pf1}
\end{equation}
Here $\mathbf{S}_t(\mathbf{k}) \in [0,1]$ serves as a phase similarity score, indicating the reliability of each frequency component.
$\odot$ denotes element-wise multiplication.
Then, $\mathbf{S}_t$ passes through a convolution layer followed by sigmoid activation produces a frequency-domain weight map:
\begin{equation}
\mathbf{W}_t = \sigma\big(\mathrm{Conv}(\mathbf{S}_t)\big), 
\quad t\in \{0,2 \}.
\label{equal:pf2}
\end{equation}

\begin{figure}[t]\footnotesize
    % \centering
    \includegraphics[width=\linewidth]{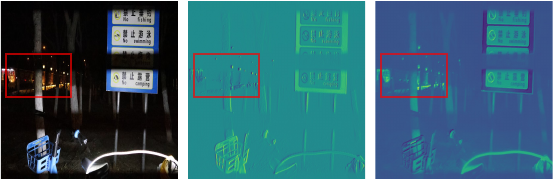}
    \begin{minipage}[t]{0.33\linewidth}
    \centering
    (a) One of the inputs
    \end{minipage}
    \begin{minipage}[t]{0.33\linewidth}
    \centering
    (b) without PFM
    \end{minipage}
    \begin{minipage}[t]{0.32\linewidth}
    \centering
    (c) with PFM
    \end{minipage}
    \vspace{-2mm}
 \caption{Feature visualization. (a) The non-reference frame provides both clean and flicker-corrupted regions to guide the base frame. (b) Without the proposed PFM, the flicker and normal regions are treated almost equally during feature fusion. (c) Benefiting from phase correlation, PFM performs an effective pre-filtering step to distinguish features in the reference frame.}
    \label{fig:attention}
\end{figure}

The dot product in the frequency domain is equivalent to convolution in the spatial domain~\cite{kong2023efficient}. Essentially, PFM leverages $\mathbf{W}_t \in \mathbb{R}^{H\times W \times C}$ as a convolution kernel to enhance the features of the reference frames. 
The enhanced frequency representations are transformed back to the spatial domain features $\hat{\mathbf{X}}_t$, which can be represented by:
\begin{equation}
\hat{\mathbf{X}}_t 
= \mathcal{F}^{-1}({\tilde{\mathbf{X}}_t \odot \mathbf{W}_t}), \quad t \in \{0,2\}.
\end{equation}
To demonstrate the effect of PFM intuitively, we visualize $\hat{\mathbf{X}}_t$ in~\cref{fig:attention}.
Finally, the enhanced spatial features are concatenated and fused together:
\begin{equation}
\mathbf{F}_0 = \text{ReLU}\big(\text{Conv}([\hat{\mathbf{X}}_0, \mathbf{X}_1, \hat{\mathbf{X}}_2])\big).
\label{equal:fusion}
\end{equation}

\begin{figure}[t]
    \centering
    \includegraphics[width=\linewidth]{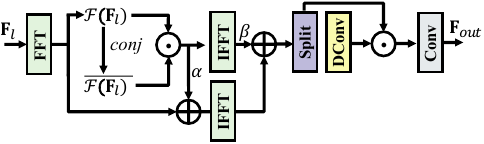}
    \caption{The mechanism of AFFN. The ``conj'' means conjugate operation. Unlike PFM, the proposed AFFN performs correlation within the same feature representation, which is referred to as autocorrelation.}
    \label{fig:AFFN}
\end{figure}

\noindent\textbf{Autocorrelation Feed Forward Network.}
Autocorrelation~\cite{auto} quantifies the similarity between a signal and its shifted versions, revealing latent periodic structures under strong noise or distortion. While PFM emphasizes inter-frame phase consistency, we further exploit intra-frame periodic cues via the proposed AFFN, as illustrated in \cref{fig:AFFN}.

To obtain the spatial autocorrelation $\mathbf{R}_l$ of the input feature map $\mathbf{F}_l \in\mathbb{R}^{H\times W\times C}$, we leverage the Wiener-Khinchin theorem~\cite{wiener}. 
The theorem states that the spatial autocorrelation $\mathbf{R}_l$ can be efficiently calculated as the inverse fast Fourier transform (IFFT) of the feature map's magnitude-squared: 
\begin{equation}
\mathbf{R}_l = \mathcal{F}^{-1}(\mathcal{F}(\mathbf{F}_l) \odot \overline{\mathcal{F}(\mathbf{F}_l)}) = \mathcal{F}^{-1} (\left | \mathcal{F}(\mathbf{F}_l) \right | ^2),
\end{equation}
where $\overline{(\cdot)}$ denotes the complex conjugation and $\left |(\cdot)\right |$ represents the magnitude in the frequency domain. $\mathcal{F}^{-1}(\cdot)$ is the IFFT operation.
The autocorrelation $\mathbf{R}_l$ amplifies repetitive spatial structures while suppressing uncorrelated noise.

To jointly leverage frequency- and spatial-domain information, we formulate a dual-domain process:
\begin{align}
\hat{\mathbf{F}}_{k} &= \mathcal{F}({\mathbf{F}_l}) + \alpha \left | \mathcal{F}({\mathbf{F}_l}) \right | ^2, \\
\hat{\mathbf{F}}_{l} &= \mathcal{F}^{-1}(\hat{\mathbf{F}}_{k}) + \beta \mathbf{R}_l ,
\end{align}
where $\alpha,\beta$ are learnable parameters balancing frequency-domain modulation and spatial-domain reinforcement.

Finally, the enhanced feature $\hat{\mathbf{F}}_l$ is processed by a depthwise gated feed-forward layer to get the output:
\begin{equation}
\mathbf{F}_{out} = \text{DWConv}\big(\text{GELU}(\hat{\mathbf{F}}^1_l) \odot \hat{\mathbf{F}}^2_l\big),
\end{equation}
where $\hat{\mathbf{F}}^1_l$ and $\hat{\mathbf{F}}^2_l$ are obtained by equal channel-wise splitting of $\hat{\mathbf{F}}_l$. Through this process, AFFN adaptively reinforces periodic regularities within the fused feature.

\subsection{Spatial-Domain Directionality Modeling}
\label{sec:WDAM}
The flicker in images aligns along the horizontal or vertical direction, which is determined by the line scanning mechanism and rolling shutter of the camera~\cite{rolling}.
Based on this directionality prior in flickering images, we propose the WDAM to enhance the sensitivity of both localized and subtle flicker artifacts.

\begin{figure}[t]\footnotesize
    \centering
    \includegraphics[width=\linewidth]{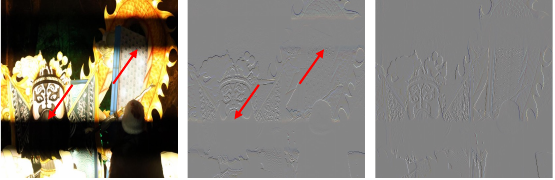}
    \begin{minipage}[t]{0.31\linewidth}
    \centering
    (a) LL
    \end{minipage}
    \begin{minipage}[t]{0.31\linewidth}
    \centering
    (b) LH
    \end{minipage}
    \begin{minipage}[t]{0.31\linewidth}
    \centering
    (c) HL
    \end{minipage}
    \vspace{-2mm}
\caption{Visualization of wavelet decomposition. The high-frequency components in LH (or HL) bands often capture regions with sharp luminance variations, which helps distinguish flickering areas from naturally dark regions and guides the attention mechanism to emphasize flicker-affected regions.}
    \label{fig:wavelet}
\end{figure}

\noindent\textbf{Wavelet-based Directional Attention.} To explicitly incorporate directionality prior into the attention mechanism, we select the Haar wavelet~\cite{huang2017wavelet} as the basis due to its inherent ability to decompose high-frequency information in horizontal and vertical directions, making it well-suited for flickering images. 
As shown in~\cref{fig:wavelet}, the edges of flicker variations are easily captured in the LH subband.
Specifically, given an input feature $\mathbf{F}_l \in \mathbb{R}^{\frac{H}{2^l} \times \frac{W}{2^l} \times 2^lC}$, we first perform a discrete wavelet transform (DWT) using the Haar basis to decompose $\mathbf{F}_l$ into one low-frequency component $\mathbf{F}_{LL} \in \mathbb{R}^{\frac{H}{2^{l+1}}\times \frac{W}{2^{l+1}} \times 2^lC}$ and three high-frequency components $\mathbf{F}_{LH}$, $\mathbf{F}_{HL}$, and $\mathbf{F}_{HH}$ with the same dimension:
\begin{equation}
[\mathbf{F}_{LL}, \mathbf{F}_{LH}, \mathbf{F}_{HL}, \mathbf{F}_{HH}] = \text{DWT}(\mathbf{F}_l),
\end{equation}
where $\mathbf{F}_{LH}$, $\mathbf{F}_{HL}$ and $\mathbf{F}_{HH}$ are horizontal, vertical and diagonal components, respectively.
The low-frequency feature $\mathbf{F}_{LL}$ is the input of the attention branch.
Following the design of window-based multi-head attention ~\cite{liu2021swin,wang2022uformer}, we split the channels into $h$ heads, each with dimensionality $d = 2^lC / h$. Then, we divide it into non-overlapping windows with size $M \times M$, obtaining a flat representation $\mathbf{F}^i_{LL} \in \mathbb{R}^{M^2 \times 2^lC}$ from the $i$-th window.
We generate \textit{queries}, \textit{keys} and \textit{values}: $\mathbf{Q}= \mathbf{F}_{LL}\mathbf{W}^Q$, $\mathbf{K} = \mathbf{F}_{LL}\mathbf{W}^K$, $\mathbf{V} = \mathbf{F}_{LL}\mathbf{W}^V \in \mathbb{R}^{\frac{HW}{4^{l}} \times 2^lC}$ using convolutions, 
where $\mathbf{W}^Q, \mathbf{W}^K, \mathbf{W}^V \in \mathbb{R}^{2^lC\times d}$ are learnable projection matrices shared by all windows.

To inject directional priors from the high-frequency subbands, we concatenate the horizontal and vertical wavelet components $\mathbf{F}_{LH}$ and $\mathbf{F}_{HL}$, and apply a $3\times3$ convolution and a sigmoid activation to generate a directional weight:
\begin{equation}
\mathbf{M} = \sigma(\text{Conv}([\mathbf{F}_{LH}, \mathbf{F}_{HL}])).
\end{equation}
The weight map $\mathbf{M} \in \mathbb{R}^{\frac{H}{2^{l+1}}\times \frac{W}{2^{l+1}}\times 2^lC}$ highlights regions where flicker artifacts are directionally dominant and serves as a learnable weighting prior for the attention mechanism.
To match the dimensionality of the value feature $\mathbf{V}$, the modulation map $\mathbf{M}$ is reshaped into a matrix of size $\mathbb{R}^{\frac{HW}{4^{l+1}} \times 2^lC}$ and divided into $h$ heads along the channel dimension to get $[\mathbf{M}_1, \mathbf{M}_2, \ldots, \mathbf{M}_h]$, where $\mathbf{M}_i \in \mathbb{R}^{\frac{HW}{4^{l+1}} \times d}, i = 1, 2, \ldots, h$.
This design ensures that each modulation sub-map $\mathbf{M}_i$ is spatially aligned with the corresponding value feature $\mathbf{V}_i$ within the same attention head.
The outputs from all heads are concatenated and projected through a linear layer to obtain the final aggregated feature.
The proposed attention mechanism is defined as:
\begin{equation}
\label{equal:att}
\textbf{Att}(\mathbf{Q}, \mathbf{K}, \mathbf{V}, \mathbf{M})
=
\text{Softmax}\left(
\frac{\mathbf{Q}\mathbf{K}^\top}{\sqrt{d}} + \mathbf{B}
\right) (\mathbf{M} \odot \mathbf{V}),
\end{equation}
where $\odot$ denotes element-wise multiplication. $\mathbf{B} \in \mathbb{R}^{\frac{HW}{4^{l+1}} \times C}$ is the learnable relative positional bias.

The refined low-frequency feature is obtained as $\mathbf{F}'_{LL}$.
The high-frequency features $\mathbf{F}'_{LH}$, $\mathbf{F}'_{HL}$, and $\mathbf{F}'_{HH}$ are generated by concatenating the original high-frequency components and passing them through a lightweight convolution.
The final output $\mathbf{F}_{att} \in \mathbb{R}^{\frac{H}{2^l} \times \frac{W}{2^l} \times 2^lC}$ is obtained by performing the inverse discrete wavelet transform (IDWT).
\noindent\textbf{Complexity Analysis.}
Let the input feature map have spatial size $H\times W$ and channel dimension $C$, and the attention window size be $M\times M$.
For standard window-based multi-head attention (W-MHA)~\cite{liu2021swin,wang2022uformer}, the computational complexity can be approximated as
\begin{equation}
\mathcal{O}_{\text{W-MHA}} = \mathcal{O}\left(\frac{HW C^2}{h} + HW M^2 C\right),
\end{equation}
In our WDAM, the attention is performed only on the low-frequency subband $\mathbf{F}_{LL}$, whose spatial size is reduced to $\frac{H}{2}\times\frac{W}{2}$ after wavelet decomposition.
The additional cost of the directional modulation map $\mathbf{M}$, generated via a $3\times3$ convolution, is linear in the number of pixels and thus negligible.
Therefore, the overall complexity of WDAM is
\begin{equation}
\mathcal{O}_{\text{WDAM}} = \frac{1}{4}\mathcal{O}_{\text{W-MHA}} + \mathcal{O}(HW C),
\end{equation}
indicating that WDAM maintains the representational power of window-based attention while reducing both computational and memory costs by approximately $75\%$.

\section{Experiments}

\begin{figure*}[t]\footnotesize
    \centering
     \includegraphics[width=\linewidth]{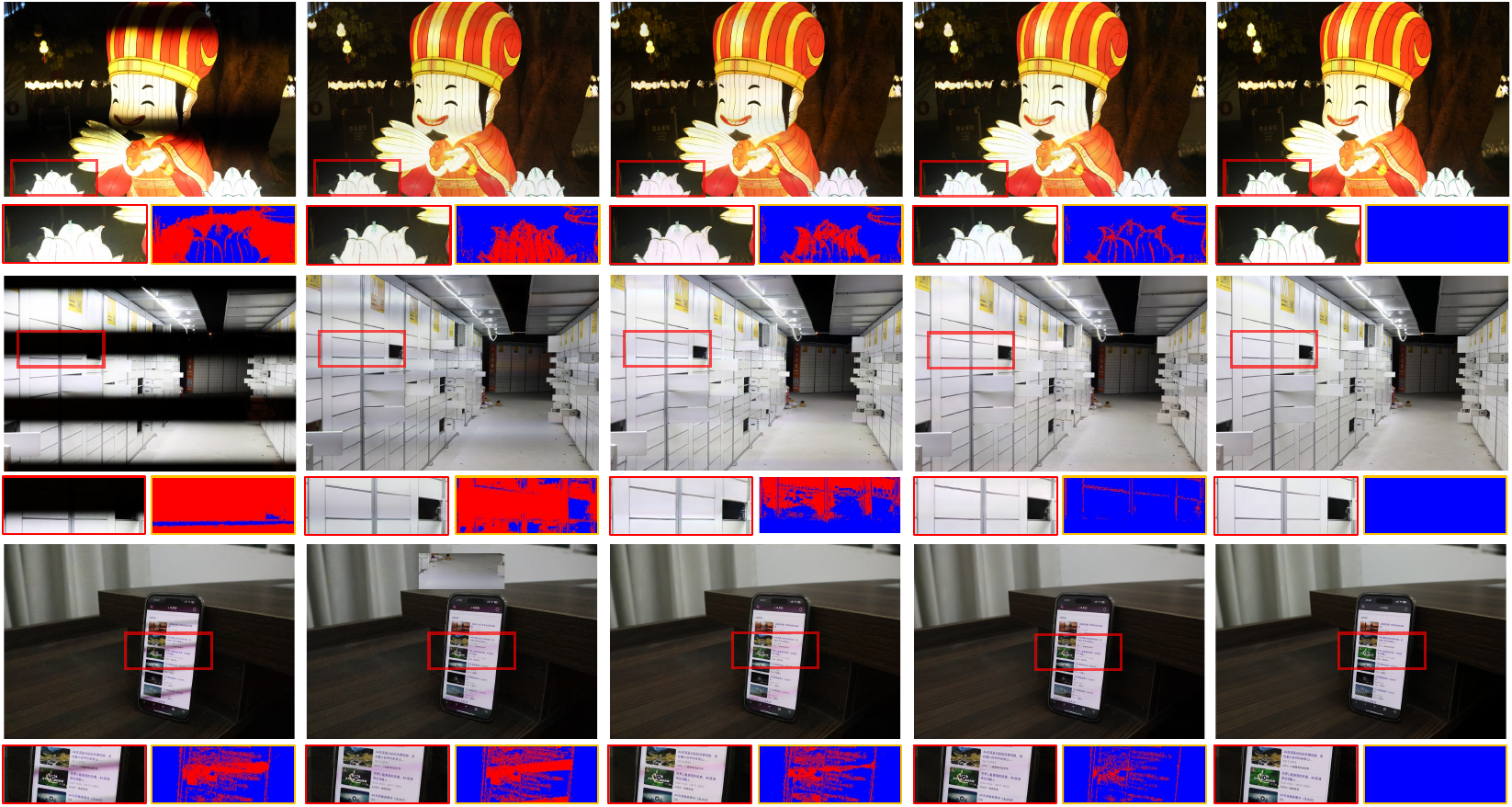}
    \begin{minipage}[t]{0.19\linewidth}
    \centering
    (a) Input
    \end{minipage}
    \begin{minipage}[t]{0.19\linewidth}
    \centering
    (b) HDRTransformer~\cite{hdrtransformer}
    \end{minipage}
    \begin{minipage}[t]{0.19\linewidth}
    \centering
    (c) AST~\cite{zhou2024adapt}
    \end{minipage}
    \begin{minipage}[t]{0.19\linewidth}
    \centering
    (d) Flickerformer (Ours)
    \end{minipage}
    \begin{minipage}[t]{0.19\linewidth}
    \centering
    (e) GT
    \end{minipage}
    \caption{Qualitative comparison of flicker removal methods. Our Flickerformer achieves superior visual quality, effectively removing flicker while preserving fine textures and color. The blue background indicates the difference map between the result and the ground truth, where pixels exceeding a certain threshold are highlighted in red.}
    \label{fig:flicker_visual}
    \vspace{-3mm}
\end{figure*}

\subsection{Experimental Settings}
To comprehensively evaluate the effectiveness of our proposed Flickerformer, we conduct experiments on the benchmark BurstDeflicker dataset~\cite{burstdeflicker}. 
Since our work is the first network specifically designed for burst flicker removal, we compare it with several state-of-the-art models originally developed for other image restoration tasks. 
All methods are trained and evaluated on the same training and testing splits to ensure fair comparison. 

\noindent\textbf{Implementation Details.} Our Flickerformer is built upon a 3-level encoder-decoder architecture. From level-1 to level-3, the number of transformer blocks is [2, 2, 2]. The number of attention heads is set to [1, 2, 4] across the levels, while the channel dimensions are set to [32, 64, 96]. Within each AFFN module, we adopt a channel expansion factor of $\gamma = 2.66$.
 The model was trained using the Adam optimizer with a learning rate of $1e^{-4}$. We employ a combined loss function with equal weights for the L1 loss and the perceptual loss using VGG-19~\cite{vggloss}. 

\begin{table}[!t]\footnotesize
\centering
\caption{Quantitative comparison of 16 methods on the benchmark dataset~\cite{burstdeflicker}. The best and second-best scores are highlighted in \textbf{bold} and \underline{underlined}, respectively.}
\vspace{-2.mm}
\scalebox{1.00}{
\footnotesize
\setlength{\tabcolsep}{0.4mm}{
\begin{tabular}{cccccccccccc}
\toprule[0.8pt]
\multicolumn{2}{l|}{\textbf{Method}}                               & \multicolumn{2}{c}{PSNR $\uparrow$}      & \multicolumn{2}{c}{SSIM $\uparrow$}          & \multicolumn{2}{c|}{LPIPS $\downarrow$}         & \multicolumn{2}{c}{Params (M)}                           & \multicolumn{2}{c}{Flops (G)}                           
\\\midrule[0.8pt]

\multicolumn{2}{l|}{Stripformer~\cite{tsai2022stripformer}}  
& \multicolumn{2}{c}{29.223}& \multicolumn{2}{c}{0.892}
& \multicolumn{2}{c|}{0.058} & \multicolumn{2}{c}{19.71} 
& \multicolumn{2}{c}{681.64}  \\

\multicolumn{2}{l|}{Uformer~\cite{wang2022uformer}}  
& \multicolumn{2}{c}{30.544}& \multicolumn{2}{c}{0.910}
& \multicolumn{2}{c|}{0.056} & \multicolumn{2}{c}{18.12} 
& \multicolumn{2}{c}{145.24}  \\

\multicolumn{2}{l|}{Restormer~\cite{zamir2022restormer}}  
& \multicolumn{2}{c}{30.630}& \multicolumn{2}{c}{0.917}
& \multicolumn{2}{c|}{0.055} & \multicolumn{2}{c}{26.10}
& \multicolumn{2}{c}{141.16}  \\

\multicolumn{2}{l|}{HDRTransformer~\cite{hdrtransformer}}   
& \multicolumn{2}{c}{30.031} & \multicolumn{2}{c}{\underline{0.918}}
& \multicolumn{2}{c|}{0.054} & \multicolumn{2}{c}{1.04}
& \multicolumn{2}{c}{272.12}  \\

\multicolumn{2}{l|}{Retinxformer~\cite{Retinexformer}}  
& \multicolumn{2}{c}{29.598}& \multicolumn{2}{c}{0.899}
& \multicolumn{2}{c|}{0.055} & \multicolumn{2}{c}{3.74} 
& \multicolumn{2}{c}{184.14}  \\

\multicolumn{2}{l|}{FFTformer~\cite{kong2023efficient}}  
& \multicolumn{2}{c}{29.478}& \multicolumn{2}{c}{0.895}
& \multicolumn{2}{c|}{0.050} & \multicolumn{2}{c}{14.88}
& \multicolumn{2}{c}{131.71}  \\

\multicolumn{2}{l|}{Burstormer~\cite{dudhane2023burstormer}}  
& \multicolumn{2}{c}{29.439}& \multicolumn{2}{c}{0.910}
& \multicolumn{2}{c|}{0.056} & \multicolumn{2}{c}{0.17} 
& \multicolumn{2}{c}{141.05}  \\

\multicolumn{2}{l|}{FBANet~\cite{wei2023towards}}  
& \multicolumn{2}{c}{29.459}& \multicolumn{2}{c}{0.896}
& \multicolumn{2}{c|}{0.052} & \multicolumn{2}{c}{4.76} 
& \multicolumn{2}{c}{432.07}  \\

\multicolumn{2}{l|}{MambaIR~\cite{zhou2025devil}}  
& \multicolumn{2}{c}{29.478}& \multicolumn{2}{c}{0.904}
& \multicolumn{2}{c|}{0.060} & \multicolumn{2}{c}{3.59}
& \multicolumn{2}{c}{186.76}  \\

\multicolumn{2}{l|}{SAFNet~\cite{kong2024safnet}} 
& \multicolumn{2}{c}{29.223}& \multicolumn{2}{c}{0.892}
& \multicolumn{2}{c|}{0.058} & \multicolumn{2}{c}{1.12} 
& \multicolumn{2}{c}{169.74}  \\

\multicolumn{2}{l|}{FPro~\cite{zhou2024seeing}}  
& \multicolumn{2}{c}{30.551}& \multicolumn{2}{c}{0.910}
& \multicolumn{2}{c|}{0.051} & \multicolumn{2}{c}{22.38}
& \multicolumn{2}{c}{247.04}  \\

\multicolumn{2}{l|}{AST~\cite{zhou2024adapt}}  
& \multicolumn{2}{c}{\underline{30.646}}& \multicolumn{2}{c}{\underline{0.918}}
& \multicolumn{2}{c|}{0.050} & \multicolumn{2}{c}{19.90}
& \multicolumn{2}{c}{156.43}  \\

\multicolumn{2}{l|}{AFUNet~\cite{li2025afunet}}  
& \multicolumn{2}{c}{28.922}& \multicolumn{2}{c}{0.903}
& \multicolumn{2}{c|}{0.066} & \multicolumn{2}{c}{1.14}
& \multicolumn{2}{c}{301.36}  \\

\multicolumn{2}{l|}{RT-XNet~\cite{jha2025rtx}}  
& \multicolumn{2}{c}{29.718}& \multicolumn{2}{c}{0.909}
& \multicolumn{2}{c|}{0.058} & \multicolumn{2}{c}{3.66}
& \multicolumn{2}{c}{245.82}  \\

\multicolumn{2}{l|}{HINT~\cite{zhou2025devil}}  
& \multicolumn{2}{c}{30.336}& \multicolumn{2}{c}{0.916}
& \multicolumn{2}{c|}{\underline{0.046}} & \multicolumn{2}{c}{24.85}
& \multicolumn{2}{c}{142.30}  \\

\midrule
\multicolumn{2}{l|}{\text{Flickerformer (Ours)}} 
& \multicolumn{2}{c}{\textbf{31.226}} & \multicolumn{2}{c}{\textbf{0.920}}
& \multicolumn{2}{c|}{\textbf{0.045}} & \multicolumn{2}{c}{3.92}
& \multicolumn{2}{c}{128.76}
\\
\bottomrule[0.8pt]
\end{tabular}}
}
\label{tab:flicker_comparison}
\vspace{-1mm}
\end{table}

\subsection{Comparisons with State-of-the-art Methods}
\noindent\textbf{Quantitative Results.} 
To provide a more comprehensive comparison, we evaluate Flickerformer against a wide range of representative models. 
Specifically, we include three networks designed for HDR reconstruction (HDRTransformer~\cite{hdrtransformer}, SAFNet~\cite{kong2024safnet}, and AFUNet~\cite{li2025afunet}), two for burst super-resolution (Burstormer~\cite{dudhane2023burstormer} and FBANet~\cite{wei2023towards}), two for deblurring (Stripformer~\cite{tsai2022stripformer} and FFTformer~\cite{kong2023efficient}), two for low-light enhancement (Retinexformer~\cite{Retinexformer} and RT-XNet~\cite{jha2025rtx}) and six general image restoration models (Uformer~\cite{wang2022uformer}, Restormer~\cite{zamir2022restormer}, FPro~\cite{zhou2024seeing}, AST~\cite{zhou2024adapt}, HINT~\cite{zhou2025devil}, and MambaIR~\cite{guo2024mambair}). 
The single-frame models in them are converted into multi-frame models by adjusting the embedding layer.
We employ three full-reference image quality metrics, including PSNR, SSIM~\cite{wang2004image}, and LPIPS~\cite{LPIPS} to assess the flicker removal performance.
Table~\ref{tab:flicker_comparison} summarizes the quantitative results of different flicker removal methods. 
Our Flickerformer achieves the best performance across all evaluation metrics. 
Specifically, Flickerformer achieves an average PSNR of 31.226 dB, outperforming the second-best method by $+0.580$ dB while using only 19.70\% of the parameters.
The superior results and lower parameter count demonstrate the effectiveness of incorporating flicker priors into the network design.

\noindent\textbf{Qualitative Results.} 
We provide visual comparisons to illustrate the visual performance of different methods in~\cref{fig:flicker_visual}. 
The variations in flickering regions can be very subtle, but they are very sensitive to the human eye when switching between frames. 
Therefore, we visualize the residual maps between the model outputs and the ground truths for a clearer presentation.
Existing methods often introduce color deviations when restoring overexposed regions (e.g., HDRTransformer~\cite{hdrtransformer} appears slightly yellow, AST~\cite{zhou2024adapt} slightly red), as shown in the first row.
For the severely flickering situation in the second row, Flickerformer achieves a more thorough restoration.
As illustrated in the third row, Flickerformer removes flicker on the screen more effectively than previous methods.
Across various flickering scenarios, our proposed Flickerformer demonstrates the best flicker localization and removal capability.

\begin{table}[!t]\footnotesize
\centering
\caption{Ablation study of using different feature refinement feed-
forward network.}
\vspace{-2mm}
\setlength{\tabcolsep}{2.5mm}{
\begin{tabular}{l|cccc|c}
\toprule
Models & \begin{tabular}[c]{@{}c@{}}FFN\\ \cite{liang2021swinir}\end{tabular} & \begin{tabular}[c]{@{}c@{}}LeFF\\ \cite{wang2022uformer}\end{tabular} & \begin{tabular}[c]{@{}c@{}}GDFN\\ \cite{zamir2022restormer}\end{tabular} & \multicolumn{1}{c|}{\begin{tabular}[c]{@{}c@{}}FRFN\\ \cite{zhou2024adapt}\end{tabular}} & \begin{tabular}[c]{@{}c@{}}AFFN \\ Ours\end{tabular} \\ 
\midrule
Params (M) & 4.45 & 4.60 & 3.92 & 4.03 & 3.92 \\
Flops (G) & 139.31 & 146.73 & 128.76 & 128.76 & 128.76 \\
PSNR (dB) & 31.876 & 30.954 & 30.959 & 30.961 & \textbf{31.226} \\
\bottomrule
\end{tabular}
}
\label{tab:ablation_ffn}
\vspace{-3mm}
\end{table}

\begin{table}[!t]\footnotesize
\centering
\caption{Ablation study of using different attention mechanisms.}
\vspace{-2mm}
\setlength{\tabcolsep}{1.2mm}{
\begin{tabular}{l|cccc|c}
\toprule
Models     & \begin{tabular}[c]{@{}c@{}}Swin SA \\ \cite{liang2021swinir}\end{tabular} & \begin{tabular}[c]{@{}c@{}}Top-k SA \\ \cite{top-k-sa}\end{tabular} & \begin{tabular}[c]{@{}c@{}}Condensed SA \\ \cite{zhao2023comprehensive}\end{tabular} & \begin{tabular}[c]{@{}c@{}}ASSA \\ \cite{zhou2024adapt}\end{tabular}  & \begin{tabular}[c]{@{}c@{}}WDAM \\ Ours\end{tabular} \\ 
\midrule
Params (M) & 3.92 & 3.96 & 3.46 & 3.92 & 3.92 \\
Flops (G)  & 139.36 & 145.20 & 132.29 & 139.42 & 128.76 \\
PSNR (dB)  & 30.896 & 30.894 & 30.981 & 30.997 &\textbf{31.226} \\
\bottomrule
\end{tabular}}
\label{tab:ablation_attention}
\vspace{-2mm}
\end{table}

\begin{figure}[t]\footnotesize
    \centering
    \includegraphics[width=\linewidth]{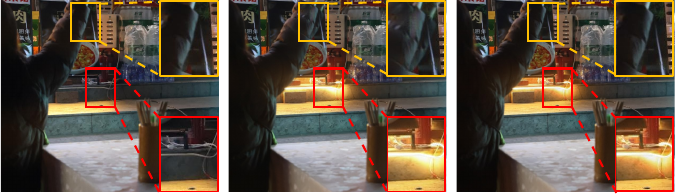}
    \begin{minipage}[t]{0.32\linewidth}
    \centering
    (a) Input
    \end{minipage}
    \begin{minipage}[t]{0.32\linewidth}
    \centering
    (b) FRFN
    \end{minipage}
    \begin{minipage}[t]{0.32\linewidth}
    \centering
    (c) AFFN (Ours)
    \end{minipage}
    \caption{Visualization comparison of different feature refinement feed-forward networks. Although both FRFN and AFFN can restore flicker-affected regions (\textcolor{red}{red} box), FRFN tends to introduce ghosting artifacts (\textcolor[RGB]{255,192,0}{yellow} box), as it struggles to distinguish between motion variations and flicker variations.}
    \label{fig:abalation_AFFN}
\end{figure}

\subsection{Ablation Study} 
We have demonstrated that Flickerformer provides favorable quantitative and visual results compared to state-of-the-art methods across various flicker scenarios. 
In this section, we present a more detailed analysis of the proposed method and the effectiveness of its key modules.

\noindent\textbf{Effect of AFFN.} 
To analyze the contribution of AFFN, we replace our AFFN with several popular alternatives, including vanilla FFN~\cite{liang2021swinir}, LeFF~\cite{wang2022uformer}, GDFN~\cite{zamir2022restormer}, and FRFN~\cite{zhou2024adapt}.
As summarized in Table~\ref{tab:ablation_ffn}, our AFFN achieves the highest PSNR while maintaining comparable model complexity and computational cost.
Specifically, compared with the FRFN~\cite{zamir2022restormer}, AFFN yields an improvement of about $+0.265$~dB in PSNR under nearly identical parameter counts.
We present visual comparisons in~\cref{fig:abalation_AFFN}, where AFFN demonstrates a strong ability to restore regions with extinguished lighting without introducing motion ghosting.

\begin{figure}[t]\footnotesize
    \centering
    \includegraphics[width=\linewidth]{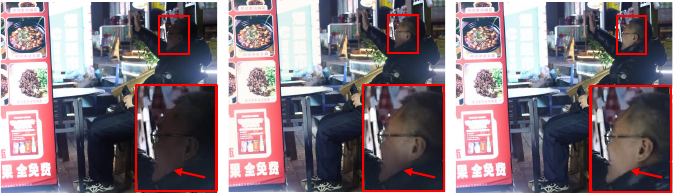}
    \begin{minipage}[t]{0.32\linewidth}
    \centering
    (a) Input
    \end{minipage}
    \begin{minipage}[t]{0.32\linewidth}
    \centering
    (b) ASSA
    \end{minipage}
    \begin{minipage}[t]{0.32\linewidth}
    \centering
    (c) WDAM (Ours)
    \end{minipage}
    \caption{Visualization comparison of different attention modules. Benefiting from directional attention and high-frequency guidance, WDAM can more accurately identify flicker-affected regions and achieve more thorough flicker removal.}
    \label{fig:abalation_WDAM}
\end{figure}

\noindent\textbf{Effect of WDAM.} 
As shown in Table~\ref{tab:ablation_attention}, replacing the proposed WDAM with conventional self-attention modules (e.g., Swin SA~\cite{liang2021swinir}, Top-k SA~\cite{top-k-sa}, Condensed SA~\cite{zhao2023comprehensive}) consistently degrades performance.
Our WDAM achieves a $+0.229$~dB gain in PSNR compared to the best alternative, with lower computational cost.
Moreover, as shown in the visual comparison in~\cref{fig:abalation_WDAM}, WDAM achieves better restoration of subtle flicker on facial regions, benefiting from its use of high-frequency information for more precise localization.

\noindent\textbf{Effect of Individual Modules.}
For the ablation study, we replace the PFM, AFFN, and WDAM modules in our Flickerformer model with their counterparts from AST~\cite{zhou2024adapt}.
As shown in~\cref{tab:abl-module}, utilizing our PFM, AFFN, and WDAM leads to PSNR improvements of $+0.279$~dB, $+0.382$~dB, and $+0.373$~dB over the AST baseline.

\begin{table}[t]
\caption{Quantitative evaluations of each module of Flickerformer. 
The CNN, FRFN, and ASSA modules are components in AST~\cite{zhou2024adapt}, while the PFM, AFFN, and MDAM are the corresponding modules in the proposed method.
}
\vspace{-2mm}
\centering
\footnotesize
\setlength{\tabcolsep}{1mm}
{
\begin{tabular}{ccccc|cccc|cccc|ccccc}
\toprule
  & \multicolumn{2}{c}{CNN}                           & \multicolumn{2}{c|}{PFM}                           & \multicolumn{2}{c}{FRFN}                        & \multicolumn{2}{c|}{AFFN}                          & \multicolumn{2}{c}{ASSA}  & \multicolumn{2}{c|}{MDAM}   & \multicolumn{2}{c}{PSNR $\uparrow$}   & \multicolumn{2}{c}{SSIM $\uparrow$} \\ 
  \midrule
(a) & \multicolumn{2}{c}{\CheckmarkBold} & \multicolumn{2}{c|}{}   & \multicolumn{2}{c}{\CheckmarkBold}                            & \multicolumn{2}{c|}{}   & \multicolumn{2}{c}{\CheckmarkBold} & \multicolumn{2}{c|}{} & \multicolumn{2}{c}{30.449} & \multicolumn{2}{c}{0.912} \\
(b) & \multicolumn{2}{c}{} & \multicolumn{2}{c|}{\CheckmarkBold}   & \multicolumn{2}{c}{\CheckmarkBold}                            & \multicolumn{2}{c|}{}   & \multicolumn{2}{c}{\CheckmarkBold} & \multicolumn{2}{c|}{} & \multicolumn{2}{c}{30.728} & \multicolumn{2}{c}{0.914} \\
(c) & \multicolumn{2}{c}{\CheckmarkBold} & \multicolumn{2}{c|}{}   & \multicolumn{2}{c}{}                            & \multicolumn{2}{c|}{\CheckmarkBold}   & \multicolumn{2}{c}{\CheckmarkBold} & \multicolumn{2}{c|}{} & \multicolumn{2}{c}{30.831} & \multicolumn{2}{c}{0.915} \\
(d) & \multicolumn{2}{c}{\CheckmarkBold} & \multicolumn{2}{c|}{}   & \multicolumn{2}{c}{\CheckmarkBold}                            & \multicolumn{2}{c|}{}   & \multicolumn{2}{c}{} & \multicolumn{2}{c|}{\CheckmarkBold} & \multicolumn{2}{c}{30.822} & \multicolumn{2}{c}{0.915} \\
\midrule
(e) & \multicolumn{2}{c}{} & \multicolumn{2}{c|}{\CheckmarkBold}   & \multicolumn{2}{c}{}                            & \multicolumn{2}{c|}{\CheckmarkBold}   & \multicolumn{2}{c}{} & \multicolumn{2}{c|}{\CheckmarkBold} & \multicolumn{2}{c}{31.226} & \multicolumn{2}{c}{0.920} \\
\bottomrule
\end{tabular}
}
\label{tab:abl-module}
\end{table}

\section{Conclusion}
In this work, we present Flickerformer, a transformer-based framework designed for flicker artifact removal by leveraging the priors of flicker degradation, namely periodicity and directionality.
To exploit the periodicity prior, we introduce two dedicated modules: the phase-based fusion module (PFM) and the autocorrelation feed-forward network (AFFN).
By leveraging the fact that phase encodes the flicker distribution, the PFM adaptively aggregates information across multiple frames through phase correlation.
The AFFN enhances recurrent structural cues after feature fusion through frequency-domain autocorrelation.
In addition, we propose the wavelet-based directional attention module (WDAM), which leverages directional high-frequency information to guide the restoration of low-frequency regions, enabling the network to capture directional dependencies and improve flicker removal performance effectively.
Extensive experiments on real-world datasets demonstrate that Flickerformer consistently surpasses state-of-the-art methods in both quantitative performance and visual quality.

\begin{figure}[t]\footnotesize
    \centering
    \includegraphics[width=\linewidth]{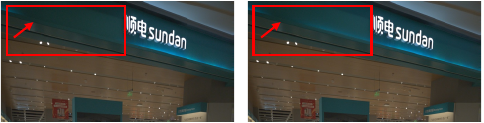}
    \begin{minipage}[t]{0.48\linewidth}
    \centering
    (a) Input
    \end{minipage}
    \begin{minipage}[t]{0.48\linewidth}
    \centering
    (b) Flickerformer (Ours)
    \end{minipage}
    \caption{Example of the limitation. Flickerformer struggles to restore regions affected by large-scale light extinction.}
    \label{fig:limitation}
\end{figure}

\noindent\textbf{Limitations.} 
When the clean regions across multiple flickering frames fail to cover the entire scene, our model struggles to restore the missing areas. 
As shown in~\cref{fig:limitation}, in the region where the long strip light turns off, only partial recovery can be achieved. 

\noindent\textbf{Acknowledgement.} This work was supported by Shenzhen Science and Technology Program (No. JCYJ20240813114229039), National Natural Science Foundation of China (No. 624B2072), Supercomputing Center of Nankai University, and OPPO Research Fund.
\clearpage
{
    \small
    \bibliographystyle{ieeenat_fullname}
    \bibliography{main}
}
\end{document}